\documentclass{article}

    \usepackage[final, nonatbib]{tackling_climate_workshop_style}

\usepackage[utf8]{inputenc} 
\usepackage[T1]{fontenc}    
\usepackage{hyperref}       
\usepackage{url}            
\usepackage{booktabs}       
\usepackage{amsfonts}       
\usepackage{nicefrac}       
\usepackage{microtype}      

\usepackage{amsmath}
\usepackage[numbers, compress, sort]{natbib}
\usepackage{graphicx}
\usepackage{multirow}
\usepackage{subcaption}
\usepackage{caption}
\usepackage{pifont}
\newcommand{\xmark}{\ding{56}}
\newcommand{\cmark}{\ding{51}}%

\title{Gaussian Processes for Monitoring Air-Quality in Kampala}

\author{%
  Clara Stoddart \\
  Dept. of Computing\\
  Imperial College London \\
  London, UK \\
  \And
  Lauren Shrack \\
  Dept. of Computer Science \\
  MIT \\
  MA, USA \\
  \And
  Richard Sserunjogi \\
  Dept. of Computer Science \\
  Makerere University / AirQo \\
  Kampala, Uganda \\
  \And
  Usman Abdul-Ganiy  \\
  Dept. of Computer Science \\
  Makerere University / AirQo \\
  Kampala, Uganda \\
  \And
  Engineer Bainomugisha \\
  Dept. of Computer Science \\
  Makerere University / AirQo \\
  Kampala, Uganda \\
  \And
  Deo Okure \\
  Dept. of Computer Science \\
  Makerere University / AirQo \\
  Kampala, Uganda \\
  \AND
  Ruth Misener \\
  Dept. of Computing\\
  Imperial College London \\
  London, UK \\
  \And
  Jose Pablo Folch \thanks{equal contribution}\\
  Dept. of Mathematics\\
  Imperial College London \\
  London, UK \\
  \And
  Ruby Sedgwick $^*$ \\
  Depts. of Materials,\\
  Bioengineering, and Computing \\
  Imperial College London \\
  London, UK \\
}
\begin{document}
\maketitle
\begin{abstract}
  Monitoring air pollution is of vital importance to the overall health of the population. Unfortunately, devices that can measure air quality can be expensive, and many cities in low and middle-income countries have to rely on a sparse allocation of them. In this paper, we investigate the use of Gaussian Processes for both $\emph{nowcasting}$ the current air-pollution in places where there are no sensors and $\emph{forecasting}$ the air-pollution in the future at the sensor locations. In particular, we focus on the city of Kampala in Uganda, using data from AirQo's network of sensors. We demonstrate the advantage of removing outliers, compare different kernel functions and additional inputs. We also compare two sparse approximations to allow for the large amounts of temporal data in the dataset.
\end{abstract}

\section{Introduction}
The World Health Organisation recognises air quality as the single biggest environmental threat to human health, estimated to cause 7 million deaths annually \cite{who}, with the negative affects of air pollution predicted to get more severe with climate change \cite{chowdhury_ambient_2018,leao_effect_2023, liu_role_2021}. Notably, particulate matter with a diameter of less than or equal to 2.5µm, known as $PM_{2.5}$, has been associated with increased all-cause, cardiovascular, respiratory and lung cancer mortality \cite{pm, gbd}. $PM_{2.5}$ has also been found to have short-term global heating effects \cite{highwood_when_2006,bond_bounding_2013, szopa_s_short-lived_2023}. 

In light of its dangerous impact, monitoring $PM_{2.5}$ levels in cities is incredibly important, but air pollution sensors are often prohibitively expensive. AirQo [\url{https://www.airqo.net/about}] are a research group based in Uganda, aiming to empower communities across Africa with information about air quality, in order to increase awareness and encourage citizens to take action against pollution. To address the current lack of air quality information in sub-Saharan Africa, AirQo has developed and deployed a sparse network of low-cost air quality monitors in multiple cities \cite{airqodata}. 

We aim to accurately predict $PM_{2.5}$ levels in Kampala, using data from AirQo's network of air quality sensors. Due to the sparse placement of the sensors, we require the predictive model to be data efficient and give a measure of uncertainty, which is useful for decision making. Gaussian processes (GPs) fulfil both these criteria, so we investigate their use for air pollution prediction, comparing the performance of a range of kernels, input factors and sparse approximations to find the best GP model for both nowcasting and forecasting. Data and code is available through the link in Appendix~\ref{app:data_avail}.

\section{Background and Related Work}

\subsection{Gaussian Processes}
A Gaussian Process \cite{rasmussen2005gps} is a stochastic process such that for any number of $n$ inputs, $x_1, ..., x_n \in \mathcal{X}$ the function $(f(x_1), ..., f(x_n))$ follows a multivariate Gaussian distribution. The distribution is defined by a mean function, $\mu_0 : \mathcal{X} \rightarrow \mathbf{R}$ and a positive definite covariance function or \textit{kernel} $\kappa_0 : \mathcal{X} \times \mathcal{X} \rightarrow \mathbf{R}$. When trying to model an unknown function, $f$, it is common to assume we have data of the form $y = f(x) + \sigma \epsilon$
where $\sigma > 0$, and $\epsilon$ is noise modelled using a standard Gaussian distribution. GP regression assumes the noise model, and puts a GP \textit{prior} on the unknown function $f$. The posterior of $f$ given a dataset $D = \{(x_i, y_i)\}_{i = 1}^N$ is also a GP whose mean and covariance can be calculated analytically. See Appendix~\ref{app:Gps} for further details.

\subsection{Kernel functions}

The kernel function specifies the covariance of the Gaussian distribution, and therefore it defines the class of functions that it can learn. We focus on the squared exponential and periodic kernels:
\begin{equation}
    \kappa_{SE}(x, x') = \exp \left(\frac{|| x - x'||^2}{2 \ell^2} \right) ; \ \kappa_{Period}(x_i, x_j) = \exp\left(-\frac{1}{2} \sum_{k=1}^{d}\left( \frac{\sin({\frac{\pi}{p}(x_{i}^{k} - x_{j}^{k}))}}{p}\right)^{2}\right) \label{eq: kernel_rbf}
\end{equation}
where $\ell$ represents the length-scale parameter, which determines how smooth the function is, and $p$ is the period parameter. The squared exponential assigns similarity to inputs that are close together, while the periodic assigns similarity between inputs that are in similar parts of the period. We can build a kernel from the product or sum of kernel functions, allowing for flexible combinations of different kernels when looking to combine different properties. For example, using a squared exponential kernel spatially and using a periodic kernel temporally. 

\subsection{Scalability of Gaussian Processes}
Computing a GP posterior requires inverting a matrix, which typically incurs cubic cost in the number of training data points, i.e. $O(N^3)$. The current state-of-the-art solution to scaling GPs is the use of sparse approximations \cite{snelson}. Fo these approximations, $M$ pseudo-inputs or \textit{inducing points} are selected such that the posterior of the GP is approximately equal to the posterior if it were to be trained in the full dataset. This results in a computational cost of order $O(M^2 N)$, where $M << N$. Spatio-temporal variational Gaussian processes \cite{hamelijnck2021spatio} provide similar results to standard variational GPs but manage to reduce the scaling in the temporal dimension from cubic to linear.

\subsection{Related Work}

GPs have shown promising results in predicting air quality \cite{reggente2014prediction, jang2020forecasting, cheng2014finegrained}. \citeauthor{reggente2014prediction} found GP regression outperformed Bayesian linear regression when forecasting particulate matter \cite{reggente2014prediction}. \citeauthor{jang2020forecasting} built on this by using a stochastic variational Gaussian process model (SVGP) \cite{hensman2013gaussian} to predict particulate concentration using data from 131 monitoring sites with hourly measurements. They found that GPs outperformed linear regression and vector auto-regressive moving-average processes \cite{jang2020forecasting}. However both these studies used other pollutants as inputs to their models, which aren't measured by AirQo's low-cost sensors so are unavailable to us. They also only focus on forecasting, where we are also interested in nowcasting.

\citet{cheng2014finegrained} develop a nowcasting GP using data from a sensor network of 51 $PM_{2.5}$ monitoring devices deployed over an area of $30km$ by $30km$. Similar to our nowcasting aim, they sought to interpolate air-quality at areas between monitoring sites, restricting their analysis to the squared exponential kernel. Their results show that GPs outperform linear interpolation and cubic spline models. 

\section{Predicting Air Pollution in Kampala}

\subsection{Dataset}

We use a dataset, provided by AirQo, of 66 air-quality sensors deployed around Kampala, using measurements from November 2021 \cite{airqodata}. The sensors provide hourly readings of $PM_{2.5}$ levels.

\begin{figure}
  \centering
  \includegraphics[width=3.7in, height=1.5in]{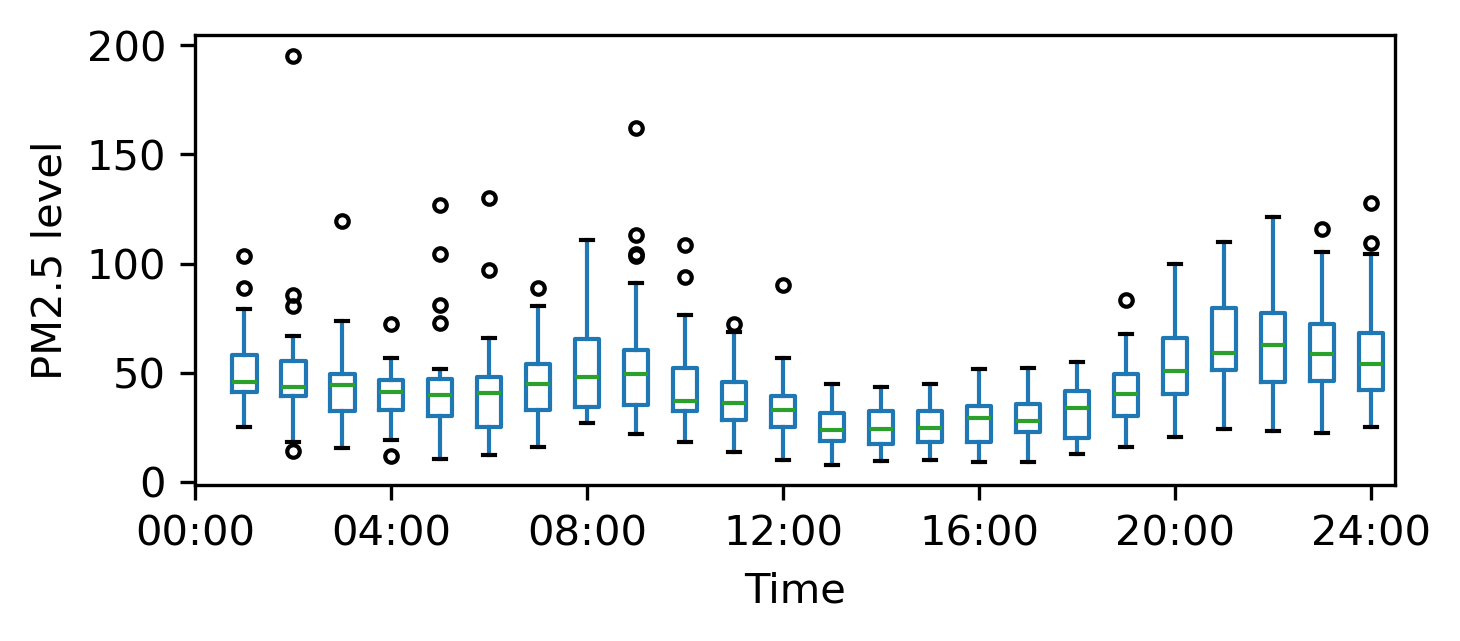}
  \caption{A box plot of $PM_{2.5}$ levels over November 2021 for the site Makerere University Weather Station. We can see a high quantity of outliers in the data; as well as a clear periodic pattern.}
  \label{fig: box_plot_makerere_with_outliers}
\end{figure}

The dataset contains a lot of extreme values, as seen in Figure \ref{fig: box_plot_makerere_with_outliers}, with many readings outside of the interquartile range, as well as many complete outliers. These outliers can be explained by sensor errors, or sudden spikes in $PM_{2.5}$ levels. These sudden spikes in pollution don't necessarily follow any spatial or temporal pattern and it is unclear if they can be predicted from the data. Indeed, spikes may occur if, for example, a car accident causes unusual levels of traffic.

There also appears to be a daily periodic pattern with higher $PM_{2.5}$ levels in the morning, at around 8am, and later in the day, at around 9pm. This pattern holds across most sites, as seen in Appendix \ref{appendix: further_data}. This motivates us to look at using a periodic kernel to predict the pollution levels.


\subsection{Building the Models}

In our experiments we investigate the effect of:
\begin{itemize}
    \item[(I)] Adding periodicity to the kernel. We do this by testing a model with a squared exponential kernel against a model with a kernel that is a sum of a squared exponential kernel applied to the spatial dimensions and a periodic kernel applied to the temporal dimensions.
    \item [(II)] Removing extreme values from the training set, under the assumption that they are caused by rare external factors that may bias the model during normal times.
    \item[(III)] Our base models only use the location and time as inputs to get predictions. We investigate the effect of including additional inputs to the model: wind speed, wind direction, wind gusts, humidity, temperature, and precipitation.
    \item[(IV)] Due to the cubic scalability of GPs, we are unable to use the whole dataset to fit standard models and use 1000 randomly selected points. We investigate the use of sparse approximations to allow the training of the model on the whole dataset.   
\end{itemize}

All models are built using the GPflow package \cite{gpflow}. For the periodic kernels, we encode a daily period and a weekly period (we do not consider monthly fluctuations as the dataset is for one month).
\subsection{Model Predictive Performance}

\begin{table}
\noindent
\caption{Comparison of different Gaussian Process models for \textbf{nowcasting}.}
\begin{tabular}{|p{4em}|p{4em}|p{4em}|p{5em}|p{5em}|p{5em}|p{5em}|}\hline
Periodic
& Outliers Removed & Additional Inputs & Sparse & Min RMSE & Average RMSE & Max RMSE \\\hline\hline
\xmark & \xmark & \xmark & \xmark &  17.12 & 27.65  & 86.36 \\\hline\hline
\cmark & \xmark & \xmark & \xmark & 11.03 & 25.49 & 87.26\\\hline\hline
\cmark & \cmark & \xmark & \xmark & 15.49 & 25.06 & 86.01\\\hline\hline
\cmark & \cmark & \cmark & \xmark & 12.98 & 24.94 & 86.50\\\hline\hline
\cmark & \cmark & \cmark & SVGP & 11.25 & 23.63 & 85.86\\\hline\hline
\xmark & \cmark & \xmark & ST-SVGP & 9.01 & 22.34 & 87.04\\\hline
\end{tabular}
\label{tab: nowcasting_results}
\end{table}

\begin{table}
\noindent
\caption{Comparison of different Gaussian Process models for \textbf{forecasting}.}
\begin{tabular}{|p{4em}|p{4em}|p{4em}|p{5em}|p{5em}|p{5em}|p{5em}|}\hline
Periodic
& Outliers Removed & Additional Inputs & Sparse & Min RMSE & Average RMSE & Max RMSE \\\hline\hline
\xmark & \xmark & \xmark & \xmark & 10.52 & 22.99 & 141.77\\\hline\hline
\cmark & \xmark & \xmark & \xmark & 9.65  & 17.75 & 44.27\\\hline\hline
\cmark & \cmark & \xmark & \xmark & 5.31  & 15.66 & 27.18\\\hline\hline
\cmark & \cmark & \cmark & \xmark & 5.89 & 15.17 & 30.15\\\hline\hline
\cmark & \cmark & \cmark & SVGP & 2.86 & 14.57 & 27.06\\\hline\hline
\xmark & \cmark & \xmark & ST-SVGP & 5.56 & 15.34 & 32.85\\\hline
\end{tabular}
\label{tab: forecasting_results}
\end{table}

For each model we investigate two different tasks. \textit{Nowcasting}, which refers to predicting pollution levels at \textit{the current time}, in an \textit{unknown location}; we test using leave-one-out cross validation on the sensor sites. 
The second task is \textit{forecasting}, where we predict the pollution levels at the \textit{current location}, in the \textit{future}; we test by holding out the last day's data. See Appedix~\ref{app:implementation_details} for implementation details. We use root mean squared error (RMSE) on the test sets as a performance metric. 

\textbf{The effect of adding periodicity} We note a very small improvement for the nowcasting task. The amount of improvement suggests that previous trends have little effect on current prediction and it is the latest measurements that are more important. For the forecasting task, however, the average RMSE notably drops across sites, with the maximum RMSE decreasing significantly.

\textbf{Removing outliers} Removing outliers has a small impact on nowcasting results, but stronger for forecasting.  It reduces the minimum, average and the maximum RMSE. The maximum RMSE is reduced significantly, explained by the aperiodcity of the spikes creating bias in the (otherwise periodic) model. We do not remove outliers from the test set for fair comparison.

\textbf{Additional variables} Wind direction, wind speed, humidity, precipitation and temperature, are included as variables into the RBF kernel, treating them the same as spatial inputs. The additional variables don't seem to lead to any clear improvement in model performance. All additional variables are measured at one weather station in Kampala; so we would hope that with more fine-grained meteorological information results might be more significant.

\textbf{Sparse approximations} are vital to allow training of GPs on full large spatio-temporal datasets. In both nowcasting and forecasting we see small improvements across all metrics.

\textbf{ST-SVGP} Spatio-temporal variational GPs \cite{hamelijnck2021spatio} reduce the temporal complexity significantly. However, we are unable to include additional inputs, as the model relies on creating a dense spatio-temporal grid.  We also lose the ability to use a periodic kernel, since it cannot easily be written in the state-space form required to use ST-SVGP. We observe improved performance for nowcasting, but worsened performance for forecasting; most likely because of the loss of periodicity.

\section{Conclusion}

In our evaluation of GP models for monitoring air quality in Kampala, we found the models to be better at forecasting than nowcasting. For forecasting, periodicity and outlier removal improved performance. For nowcasting, the maximum RMSE was similar amongst all nowcasting models, suggesting difficulty in predicting certain locations without more sensors and perhaps the need for a non-stationary kernel \cite{patel_accurate_2022}. We hope this work can help guide pollution predictive models and sensor placement in the future \cite{krause2008near}. 


\clearpage

\bibliography{references}

\clearpage

\appendix

\section{Acknowledgments}

We would firstly like to acknowledge the AI for Social Good seminar at Dagstuhl (seminar no. 22091) where the collaboration began, in particular we would like to thank Raghu Rajan and Asma Atamna for valuable feedback in previous iterations of the project. Richard S, Usman AG, Engineer B and Deo O received support from Belgium through the Wehubit programme implemented by Enabel and from Google LLC. Jose Pablo F is funded by EPSRC through the Modern Statistics and Statistical Machine Learning (StatML) CDT (grant no. EP/S023151/1) and by BASF SE, Ludwigshafen am Rhein. Ruby S was supported by the UKRI AI for Healthcare CDT (grant no. EP/S023283/1).

\section{Data Availability} \label{app:data_avail}

The full code and dataset will be available at \url{https://github.com/claramst/gps-kampala-airquality}.

\section{Further Background on Gaussian Processes} \label{app:Gps}

A Gaussian Process \cite{rasmussen2005gps} is defined as a stochastic process whose finite dimensional distributions are jointly Gaussian. That is, for some input space $\mathcal{X}$, and any number of $n$ inputs, $x_1, ..., x_n \in \mathcal{X}$ then $(f(x_1), ..., f(x_n))$ follows a multivariate Gaussian distribution. The distribution can be completely defined by a mean function, $\mu_0 : \mathcal{X} \rightarrow \mathbf{R}$ and a positive definite covariance function or \textit{kernel} $\kappa_0 : \mathcal{X} \times \mathcal{X} \rightarrow \mathbf{R}$.

When trying to model an unknown function, $f$, it is common to assume we have data of the form:
\begin{equation}
    y = f(x) + \sigma \epsilon
\end{equation}
where $\sigma > 0$, and $\epsilon$ is noise modelled using a standard Gaussian distribution. Gaussian Processes regression assumes the model above, and puts a Gaussian Process \textit{prior} on the unknown function $f$. It can then be shown that the posterior of $f$ given a dataset $D = \{(x_i, y_i)\}_{i = 1}^N$ is also a Gaussian Process, with mean and covariance functions given by:
\begin{align}
    \mu(\mathbf{f}_{*}) &= K(X_*,X)[K(X, X) + \sigma^2 I]^{-1}(\mathbf{y} - \mu_0(X)) \label{eq: gp_posterior_mean} \\
    \sigma(\mathbf{f}_{*}) &= K(X_{*}, X_{*}) - K(X_*,X)[K(X, X) + \sigma^2 I]^{-1} K(X, X_{*}) \label{(eq: gp_posterior_covariance}
\end{align}
where $X_*$ is a matrix of target locations, $\mathbf{f}_{*}$ the corresponding target outputs, $X$ is the training inputs, $\mathbf{y}$ is the training observations, $[\mu_0(X)]_i = \mu_0(X_i)$ and $[K(X, X')]_{i, j} = \kappa_0 (x_i, x'_j)$. The hyper-parameters of the kernel and the noise level in the likelihood can all be learnt from data by maximizing the log marginal likelihood of the model.

The matrix inversion in equations (\ref{eq: gp_posterior_mean}) and (\ref{(eq: gp_posterior_covariance}) lead to the cubic scaling in GPs. When using sparse approximations to alleviate the problem, it is common to treat the pseudo-inputs as variational parameters \cite{titsias} and learn them along the kernel hyper-parameters.

\section{Further Data Description and Analysis} \label{appendix: further_data}

Low-cost air quality monitors are prone to errors as they may be affected by weather conditions, and their overall accuracy degrades over time. AirQo use machine learning models (random forests and lasso regression) to calibrate the data in order to mitigate against errors \citep{adong2022applying, smith2022modelling}. We use the calibrated $PM_{2.5}$ levels after AirQo's processing. Some sensors have a number of missing or null readings due to errors, meaning that not all sites have the same number of readings for a given time frame. The full dataset for November 2021 contained readings from 68 sites across Kampala, however, 2 of the sites had less than 100 readings across the whole month so they were removed. 

\begin{figure}
  \centering
  \includegraphics[width=5in, height=2in]{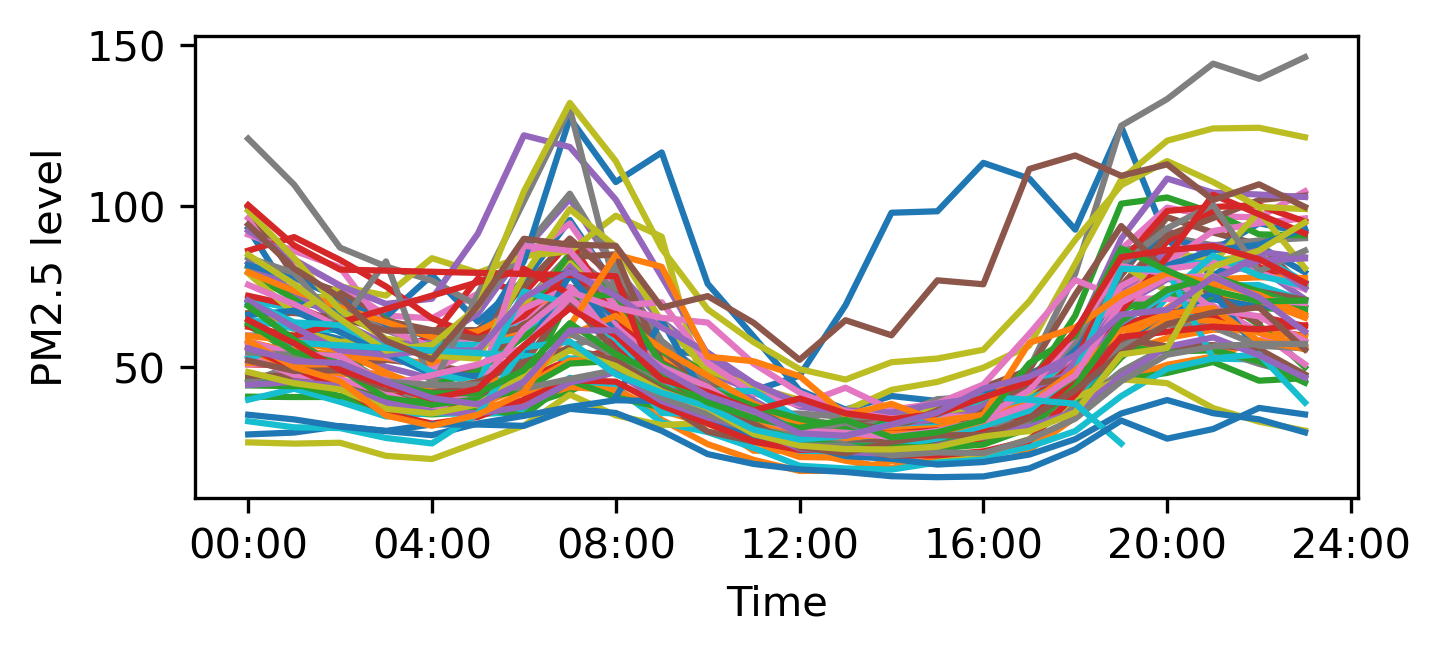}
  \caption{Mean $PM_{2.5}$ levels at each time for the 66 sites in Kampala over September 2021. We can see clear periodic patterns related to the daily commute of people in the city.}
  \label{fig: data_analysis}
\end{figure}

Figure \ref{fig: data_analysis} displays the mean $PM_{2.5}$ levels for each of the 66 sensors across the city. We can observe the clear periodic pattern, with pollution peaks around 8am and 9pm.

\section{Implementation Details} \label{app:implementation_details}

For the standard GP, we use an RBF kernel. Due to the size of the dataset, we cannot train using the full dataset. Therefore we randomly select 1000 training points before fitting non-sparse GPs; due to the randomness in the process we repeat each experiment 4 times and take an average -- we do this for all non-sparse methods. We built all models using GPFlow \cite{matthews2017gpflow}, with the exception of spatio-temporal variational Gaussian processes (ST-SVGP) \cite{hamelijnck2021spatio} where we used the implementation provided by the authors.

\textbf{The effect of adding periodicity} is tested by using a linear combination of an RBF kernel on the spatial dimensions and a periodic kernel on temporal ones. The periodic part is created using a product of a periodic kernel with a daily period and one with a weekly period (we do not consider monthly fluctuations as the dataset is for one month only). 

\textbf{Removing outliers} We remove observations that are over 1.5 times the inter-quartile range away from the mean and comparing model performance. We do not remove outliers from the test set for fair comparison. 

\textbf{Additional variables} are treated as spatial variables in an RBF kernel. There are publicly accessible sources of this information, and so we used data from the Visual Crossing Weather data service \cite{visualcrossing}. We do note that this data is sourced from only one weather data station in Kampala, and we take this information to be accurate across the whole city. In reality, wind speed may be slightly different on the other side of the city.

\textbf{Sparse approximations} A vital decision when using sparse approximations is the number of inducing points to use -- the more inducing points the more accurate the approximation, but the computational cost increases as well. For the standard sparse variational approximation, \textbf{SVGP}, we decided on 100 inducing points. We selected this after seeing diminishing gains for nowcasting (we tested on 10, 50, 100, and 200 points). This allowed the sparse model to train on the full dataset in approximately two hours. For this model we included additional variables.

\textbf{ST-SVGP} We were able to comfortably use 66 inducing points, which effectively corresponds to using 66 $\times$ 720 inducing points in the SVGP model. Additional variables could be included as spatial ones, however, this would cause the spatio-temporal grid to become too large and make the method computationally infeasible.

\end{document}